**DETC2006-99065.**

# A SIX DEGREE-OF-FREEDOM HAPTIC DEVICE BASED ON THE ORTHOGLIDE AND A HYBRID AGILE EYE

**Damien Chablat**
*Institut de Recherche en Communications et Cybernétique de Nantes*
*1, rue de la Noë, 44321 Nantes, France*
Damien.Chablat@irccyn.ec-nantes.fr

**Philippe Wenger**
*Institut de Recherche en Communications et Cybernétique de Nantes*
*1, rue de la Noë, 44321 Nantes, France*
Philippe.Wenger@irccyn.ec-nantes.fr

**ABSTRACT**

This paper is devoted to the kinematic design of a new six degree-of-freedom haptic device using two parallel mechanisms. The first one, called orthoglide, provides the translation motions and the second one, called agile eye, produces the rotational motions. These two motions are decoupled to simplify the direct and inverse kinematics, as it is needed for real-time control. To reduce the inertial load, the motors are fixed on the base and a transmission with two universal joints is used to transmit the rotational motions from the base to the end-effector. Two alternative wrists are proposed (i), the agile eye with three degrees of freedom or (ii) a hybrid wrist made by the assembly of a two-dof agile eye with a rotary motor. The last one is optimized to increase its stiffness and to decrease the number of moving parts.

## 1 INTRODUCTION

The aim of this paper is the kinematic design of a new haptic device that can be used as a new interface in a Virtual Reality system. Virtual Reality, characterized by real-time simulation, allows a human to interact with a Virtual 3D world and to understand its behavior by sensory feedback [1]. Associated with this definition, four key elements are usually considered: (i) virtual world, (ii) immersion (of human), (iii) interactive, and (iv) sensory feedback. Most often, Virtual Reality systems provide only visual sensory feedback to the user (mono or stereo vision). However, aural or haptic interfaces exist that allow the user to hear or to feel the virtual environment. We want our device to be able to feel the virtual environment.

The haptic feedback is associated with the sense of touch. For a user, the existence of an object in a virtual world is verified by coming into physical contact or touch. Usually, a haptic device enables input and output communication between the computer and the user.

Haptic devices play important roles in the recognition of virtual objects. With these devices, the user can feel the rigidity or weight of virtual objects. Most robotic manipulators have large-scale and high-cost hardware, which inhibits their application to human-computer interaction.

Our device was specifically developed for desktop use. It provides haptic feedback, which strongly enhances human capabilities in the major application areas of virtual reality, such as scientific visualization and 3D shape modeling.

The six-dof Orthoglide features two parallel mechanisms mounted serially with a suitable system, which makes it possible to know and control the orientation of a stylus, even though all the motors are fixed. These features contribute to a significant reduction of the inertia when compared to serial architectures commonly used in haptic devices [2]. Thus, the inertia of the manipulator's moving parts is so small that compensation is not needed. This hardware design is compact, and it has the ability to carry a relatively large payload. Furthermore, the parallel kinematic architecture increases drastically the stiffness of the structure.

## 2 STATE OF THE ART OF HAPTIC DEVICES

Basically, haptic devices that enable the user to touch an object can be classified into two main types: the first one provides only one single point of contact and the second one provides a single point of contact with a torque or multiple points of contact.

### 2.1 Single point of contact

Many haptic devices have only three degrees of freedom, such as the Phantom Desktop [2] depicted in Fig. 1. With such a device, the user drives a stylus in space and can feel a single point of contact with an object. The force feedback provides stimuli to fingertip but no torque is provided.

    

To provide high stiffness and a low inertia, parallel mechanisms can be also experimented as depicted in Fig. 2. This design has been optimized to avoid the singular configuration inside the workspace by Leguay-Durand and Reboulet [3].

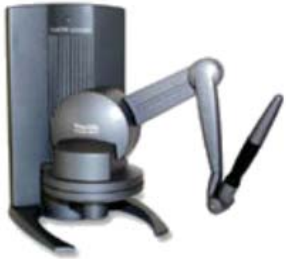

**Figure 1: The Phantom Desktop provides one single point of contact to the user**

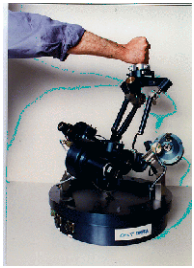

**Figure 2: The "syntaxeur" from ONERA/CERT [3]**

## 2.2 Single point of contact with torque or multiple points of contact

With a three-dof haptic device, the user can only manage the motion of one point. However, in virtual environment, he may also want to handle objects. For example, when one assembles a screw in a hole, it is necessary to produce a feedback by forces and torques to feel the contact. This feature is related to the problem of grasping.

Only a six-dof device can provide forces and torques at the same time and there are three types of mechanisms:

-Serial: The main problem of the serial type is the lack of stiffness. Therefore, the position and the rotation are coupled. An example of the first type, depicted in Fig. 3(a), was originally designed at McGill University by Hayward [4] and is now commercialized by MPB [5]. The Phantom Premium six-dof device is also of the serial type but with a closed loop in the serial chain.

-Parallel: An example of this type, depicted in Fig. 3(b), was designed at Tsukuba University [6]. The main advantage of this device is that all the actuators are fixed on the base. However, its rotation motions are limited and are function of the position. The computation of the inverse and direct kinematic models is not simple, which is not compatible with real time applications.

-Hybrid: An example of this type, depicted in Fig. 3(c), was designed at École Fédérale de Lausanne and is now commercialized by Force Dimension [7]. A parallel architecture provides high stiffness but the shape of its Cartesian workspace is generally not simple and few of them have an isotropic configuration.

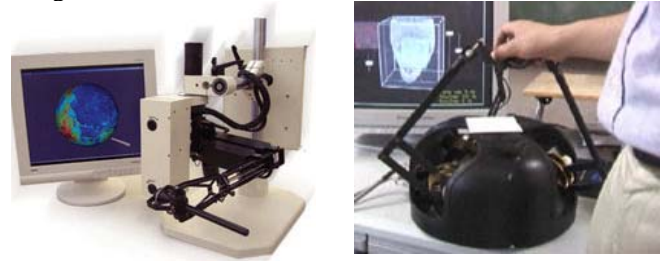

(a) (b)

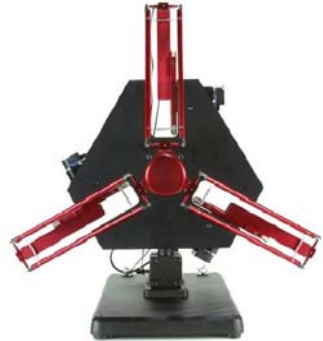

(c)

**Figure 3: Three haptic devices with six-dof with (a) serial architecture, (b) parallel architecture and (c) hybrid architecture**

Figure 4 depicts a haptic device created by Tsumaki, et al. [8] made of a Delta [9] a two-dof agile eye [10] mounted serially with a revolute joint. Such a design enables to decouple the translation motions and the rotation motions. The kinematics models are simple but its main drawbacks are the actuated joints of the wrist, which increase the inertia in motion and its volume. The aim of our paper is to propose a design with the same advantage but without these drawbacks.

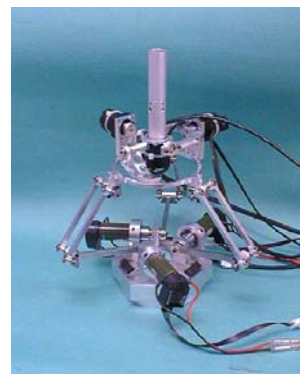

**Figure 4: A six-dof haptic device from Tohoku University**

    

# 3 CONSTRAINTS TO DESIGN A NEW HAPTIC INTERFACE

The main desirable characteristics are quick motion and a regular workspace shape. In addition, compactness is also desirable, since the haptic interface should be treated in the same way as a conventional computer mouse. Quick motions could be achieved with a parallel mechanism since the mass of the end-effecter can be reduced [11].

It is possible to build a compact haptic device with a six-dof parallel mechanism, as shown by Long, et al. [12].. In general, fully six-dof parallel mechanisms (e.g. Stewart Platform [13], HEXA [14]) have a restricted orientation workspace. To cope with this problem, it is possible to mount a spherical wrist on a translating mechanism.

Without a force/torque sensor in the haptic interface, a fully isotropic mechanism would be needed to ensure uniform force controllability in each direction. Unfortunately, six-dof isotropic mechanisms do not exist because isotropy cannot be defined due to the non-homogeneity of the Jacobian matrix as is shown by Angeles [16]. Conversely, using a force/torque sensor would increase the inertia and the volume of the device.

In general, the size of the workspace of haptic devices is relatively small compared to the size of the simulated virtual environment. Thus, there are several problems relating to the correspondence between displacements in these two spaces as scale factor, reachable space and directions of the translations or senses of rotations. To simplify these problems, the shape of the workspace of haptic devices must be close to a cube because its projection of the plan of view is the screen of the computer.

To summarize these constraints, we can define an ideal haptic interface with the following constraints:

- To be able to display motions in 6-dof,
- To be able to realize quick motion (low inertia),
- To have a regular dextrous workspace like a cube
- To be compact

Other design constraints, taking into account the control and the software can be found in Hayward [17].

## 3.1 Fully isotropic mechanisms

In the same times, a three-dof fully isotropic parallel mechanism was defined by Kong and Gosselin (2002), Kim and Tsai (2002) and Carricato and Parenti-Castelli (2002) [18, 19, 20]. This device was firstly designed for milling applications. The linear joints can be actuated by means of linear motors or by conventional rotary motors with ball screws. The output body is connected to the linear joints through a set of three *RRR* identical chains where *R* stands Revolute joints.

Figure 5 depicts an example of such a mechanism where the three prismatic joints ① are orthogonal, and the three joint axes ②, ③ and ④, are parallel. Its workspace is defined by the intersection of three cylinders and no singular configurations exist inside.

We could use this mechanism as a haptic device with three-dof because full isotropy is clearly an outstanding property. Unfortunately, bulky legs are required to assure stiffness because these legs are subject to bending. To increase the stiffness means to increase the inertia. Moreover, it would very difficult to mount a wrist on this mechanism and to actuate if from fixed motors. Another solution would be to add one or two legs as is made by Gogu [21] or by Chablat and Wenger [23] but mechanical interferences would reduce the size of the workspace.

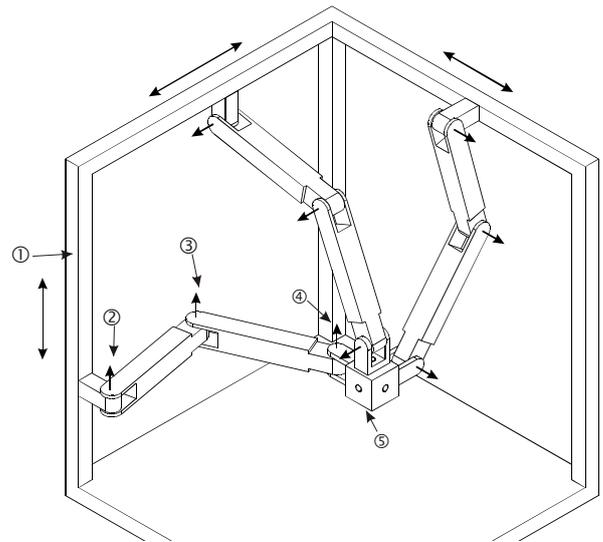

**Figure 5: A fully isotropic parallel mechanism**

## 3.2 Orthoglide

The Orthoglide is part of the Delta linear family. It was optimized to have an isotropic configuration. Thus, its leg lengths as well as its range limits were also defined to have homogeneous performances throughout its Cartesian workspace [23]. The criterion used was related the Jacobian matrix and made it possible to characterize the velocity factor amplifications. The resulting machine, the Orthoglide, features three fixed parallel linear joints, which are mounted orthogonally and a mobile platform, which moves in the Cartesian *x-y-z* space with fixed orientation. The interesting features of the Orthoglide are a regular Cartesian workspace shape and good compactness.

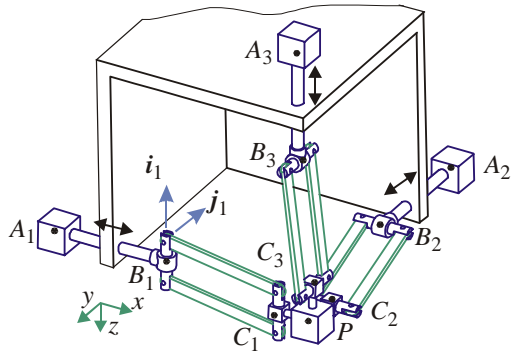

**Figure 6: Orthoglide**

The three legs are *PRPaR* identical chains, where *P*, *R* and *Pa* stands for Prismatic, Revolute and Parallelogram joint, respectively. The arrangement of the joints in the *PRPaR* chains was defined to eliminate any special singularity [24]. Each base point $A_i$ is fixed on the $i^{th}$ linear axis such that $A_1A_2 = A_1A_3 = A_2A_3$. The points $B_i$ and $C_i$ are located on the $i^{th}$ parallelogram as shown in Fig. 7 (a).

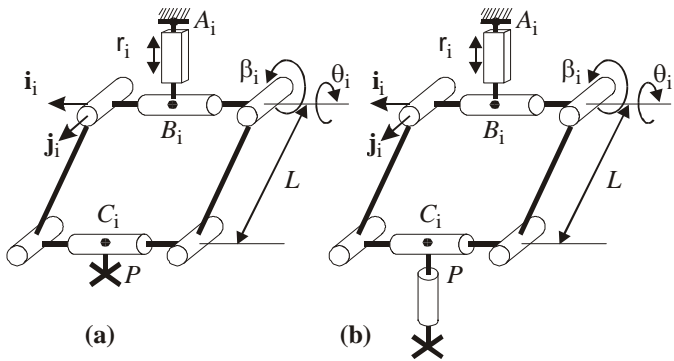

**Figure 7: Legs kinematics of the Orthoglide for the overconstraint and non-overconstraint mechanism**

Figure 7 (b) depicts another leg kinematics where a revolute joint is added to have a non-overconstraint mechanism [25]. This way, the constraints in the leg are only traction and compression type. This feature permits us to use composite material for the leg and reduce the masses.

### 3.3 Wrist mechanism

The architecture of the wrist can be either serial or parallel. In some haptic devices, the actuators are fixed to the base and the motion is transmitted to the end-effector by using cable. In other cases, the motors are embedded.

To have high stiffness, the parallel architectures are suitable but it suffers from a limited workspace. An exhaustive study of such mechanisms can be found in [26]. Some of them are able to have a fixed center point in the same loci of the center of the user hand. Moreover, particular designs can provide unlimited rotations about this central point, allowing very large workspace, limited only by the low-scaled dexterity and mechanical interferences. Figure 8 depicts an example of such a device made at Laval University [27].

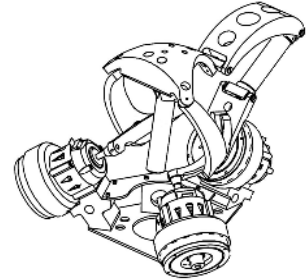

**Figure 8: The ShaDe device**

A parallel wrist with actuation redundancy can be employed to avoid the singularities [15]. An example of such a device is depicted in Fig. 9. But it would be very difficult to actuate this device with four motors fixed on the base of a translating mechanism.

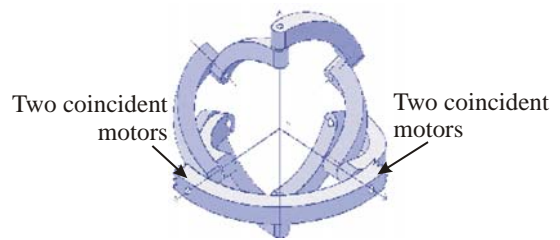

**Figure 9: Redundant spherical wrist**

An hybrid architecture derived from the two-dof agile eye is depicted in Fig. 10. It is the same one that is used Tsumaki, et al. [8]. Its main property is that there are unlimited motions about its last joint. Unfortunately, there is no way to place the last motors on the base.

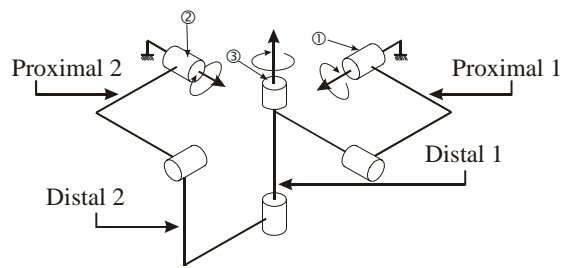

**Figure 10: The hybrid agile eye with two-dof in serial with a revolute actuated joint ③**

The kinematic behavior of this mechanism is good because there is a set of orientations where the Jacobian matrix is isotropic as shown in Fig. 10. In the home posture, the joint axis ③ is orthogonal to the joint axes ① and ②.

   

### 3.4 Design of the legs

The aim of our design is to find a mechanism that allows us to avoid to carry the motors of the wrist and to reduce its volume. We have chosen to use the agile eye with two or three dofs because with such a mechanism we can easily achieve the orientation range needed for haptic devices. We can also notice that this design can be simplified to have a six-dof mechanism with only three-dof haptic feedback.
The kinematics of the new leg is depicted in Fig. 11.

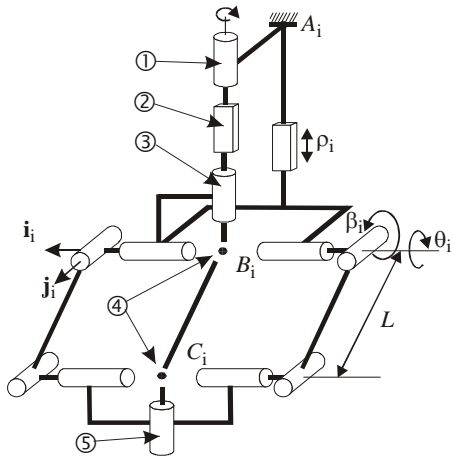

**Figure 11: New design for the legs of the Orthoglide**

On the neutral fiber of a leg, a mechanism with two universal joints is used to transmit the rotary motion from the base of the leg to the spherical wrist. From the kinematics of the Orthoglide depicted in Fig. 7, we have added a revolute actuated joint ① fixed on the base followed by a prismatic joint and a revolute joint, called ② and ③, respectively. This former is carried by the prismatic actuated joints. Two universal joints ④, located on $B_i$ and $C_i$, are placed on the neutral fiber of the parallelogram and finally, a revolute joint ⑤ allows us to mount the agile eye on the previous end-effector of the Orthoglide. To have a three-dof agile eye, all the legs are changed and to have a two-dof agile eye, only two are.

### 3.5 New haptic devices

The main problem to design a six-dof haptic device is to couple the wrist mechanism and the position mechanism. With the leg design, we can produce two types of mechanisms using either a three-dof parallel spherical wrist or a hybrid spherical wrist based on the two-dof agile eye. We use the fact that on connection point of the three legs, the joint axes ⑤ in Fig. 11 are orthogonal. Such a feature allows us to assemble the two versions of the agile eye for which there are isotropic configurations.
To compare the two designs, it is necessary to recall the shape of the Cartesian workspace of the Orthoglide as it is depicted in Fig. 12. The largest cubic workspace is reached whenever the faces of the cube are parallel to the *xy*, *yz* and *zx* plane, respectively.

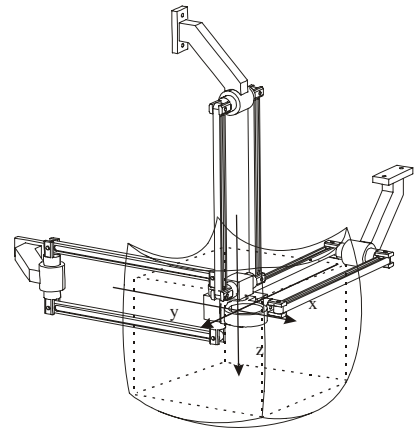

**Figure 12: The workspace of the orthoglide and the largest cubic workspace included inside**

Figures 13 and 14 depict two new haptic devices using the Orthoglide and the 3-dof (resp. 2-dof) agile eye. The two architectures have been first designed for milling machine applications by Chablat and Wenger [28].

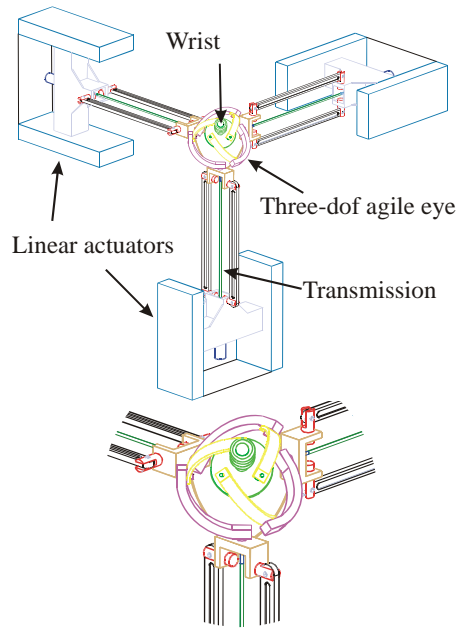

**Figure 13: A six-dof haptic device 3T+3R**

In figures 13, no revolute actuated joints are embedded. An isotropic configuration exists whenever the three legs are orthogonal and the last axis of the wrist is along the axis equal to $x = y = z$, *i.e.* on the diagonal of the cubic workspace defined in Fig. 12. This feature can be a drawback for the user because the home posture of the wrist does not lie on the axis of the cubic workspace. For the agile eye, the shape of the moving parts is complex if we want to have a high stiffness as is shown in Bidault, et al. (2001). Thus, its mobility is limited by singular configurations and internal collisions.

To avoid these drawbacks, in Fig. 14, we propose a second mechanism, which is made of the assembly of the orthoglide, the two-dof agile eye and a revolute actuated joint (3T+2R+1R).

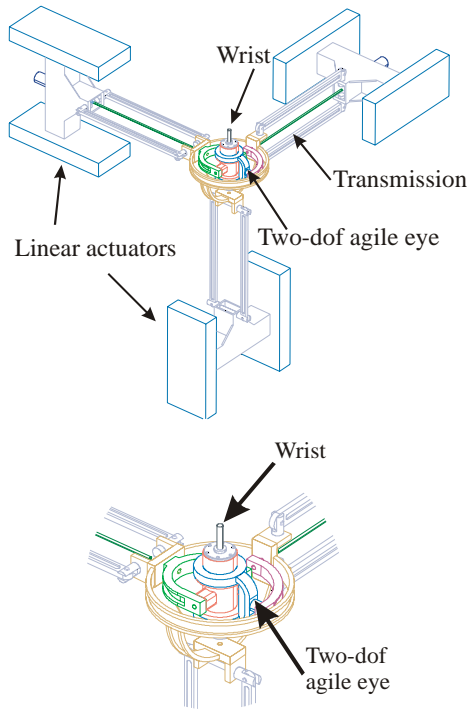

**Figure 14: Six-dof haptic device 3T+2R+1R**

This device has an isotropic configuration for the translation and the rotation part whenever the three legs are orthogonal and the last axis of the wrist is collinear with *z*-direction. In this case, one revolute actuated joint is embedded. If this feature increases the inertia, it provides unlimited motions about this axis. The wrist is able to produce pitching and yaw equal to ± 45 degrees and unlimited rolling.

### *3.6    A new design of the two-dof agile eye*

The two-dof agile eye was firstly design to carry a camera or a laser beam. For a milling machine as well as a haptic device, we need to have more stiffness. To do this, we keep the kinematics given in Fig. (10) and we change the shape of the pieces. In fact, to increase the stiffness, we design the two proximal ① and ② such that they are posing on two supports to limit their own deflection. On the first design made by Gosselin and Hamel (1994), such property cannot be achieved due to interference between the links.

However, if we change the position of the joint between the proximal and distal ②, we can produce a new design for the proximal ① with two revolute joints in contact with the base and two revolution joints in contact with the distal ① as shown in Fig. (15).

The proximal ② feature also two revolute joints in contact with the base. Thus, the size of the distal ② is reduced because it slides in a groove made in the proximal ②. A revolute joint is located between proximal ① and ②.

The wrist can also be non-overconstraint if we change the type of joint between proximal and distal ②. In fact, a simple contact point is required in the groove of the proximal ②.

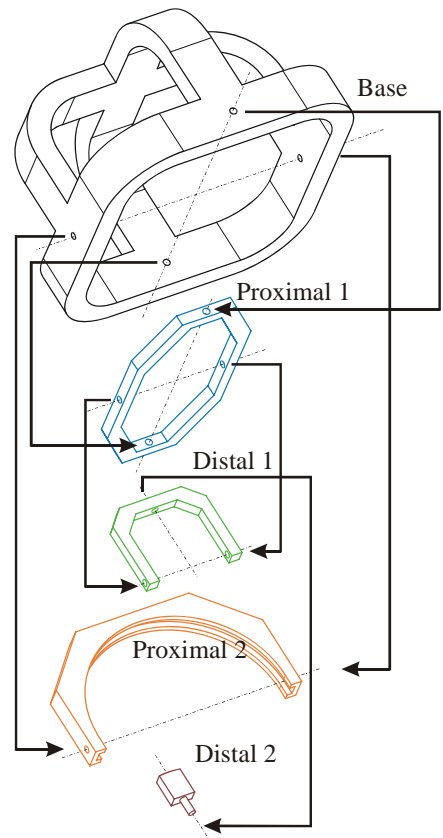

**Figure 15: New part design of the agile eye**

Figures 16 and 17 depict the assembly of the proximal and distal parts on the base with an embedded motors which control the orientation of the third joints of the wrist. This motor is connected with the distal ① and its axis is directly connected with the hand of the human.

With this design, we have no interference between the proximal and distal parts. The mechanism can rotate freely from its isotropic configuration to the singular configurations, which are reached whenever the first or the second joint values are −90 deg or +90 deg.

   

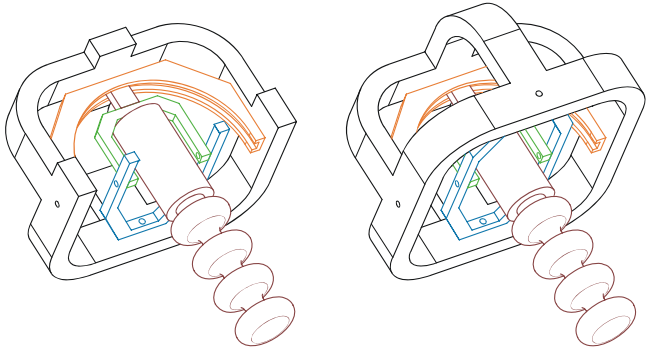

**Figure 16: New design of the wrist**

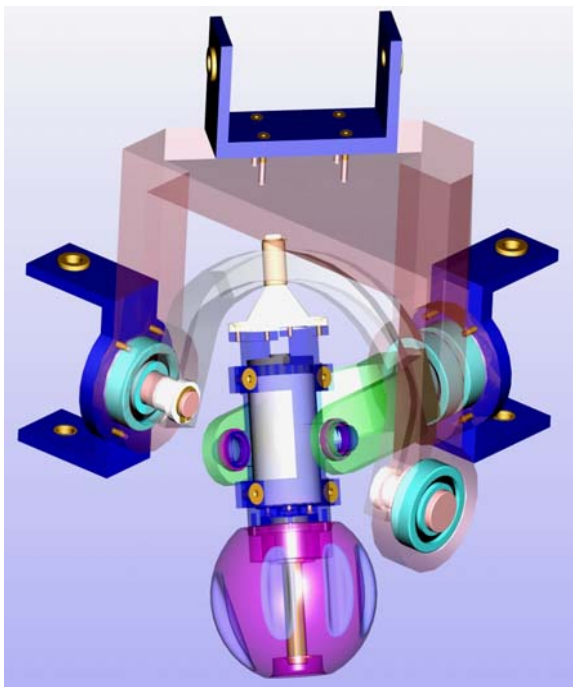

**Figure 17: CAD model of the new wrist**

## 4 CONCLUSIONS

In this paper, two haptic devices with six-dof have been defined. The first one uses two fully parallel mechanisms (3T+3R) and the second one two fully parallel mechanism (3T+2R+1R) in serial with a terminal joint. In both cases, a specific transmission is used to avoid carrying the motors of the wrist and to reduce its volume. The volume and the shape of the translation workspace were compared with the rotation workspace to have the best solution. Two architectures of wrist were proposed without and with embedded motors. The most suitable solution seems to be the second solution because its kinematics is simpler and feature fewer moving parts. A new design of this hybrid wrist was made to increase its stiffness.